\documentclass[10pt,twocolumn,letterpaper]{article}

\usepackage{cvpr}
\usepackage{times}
\usepackage{epsfig}
\usepackage{graphicx}
\usepackage{amsmath}
\usepackage{amssymb}

\cvprfinalcopy 

\def\cvprPaperID{3875} 

\ifcvprfinal\fi
\begin{document}

\title{Face-Focused Cross-Stream Network for Deception Detection in Videos}

\author{Mingyu Ding$^{1}$~~~An Zhao$^{1}$~~~Zhiwu Lu$^{1}$\thanks{Corresponding author.}~~~Tao Xiang$^{2}$~~~Ji-Rong Wen$^{1}$\\
  $^1$School of Information, Renmin University of China, Beijing 100872, China\\
  $^2$School of EECS, Queen Mary University of London, London E1 4NS, U.K.\\
  {\tt\small zhiwu.lu@gmail.com~~~~t.xiang@qmul.ac.uk} \\
}

\maketitle

\begin{abstract}
 Automated deception detection (ADD) from real-life videos is a challenging task. It specifically needs to address two problems:  (1) Both face and body contain useful cues regarding whether a subject is deceptive. How to effectively fuse the two is thus key to the effectiveness of an ADD model. (2) Real-life deceptive samples are hard to collect; learning with limited training data thus challenges most deep learning based ADD models. In this work, both problems are addressed. Specifically, for face-body multimodal learning, a novel face-focused cross-stream network (FFCSN) is proposed. It differs significantly from the popular two-stream networks in that: (a) face detection is added into the spatial stream to capture the facial expressions explicitly, and (b) correlation learning is performed across the spatial and temporal streams for joint deep feature learning across both face and body. To address the training data scarcity problem, our FFCSN model is trained with both meta learning and adversarial learning. Extensive experiments show that our FFCSN model achieves state-of-the-art results. Further, the proposed FFCSN model as well as its robust training strategy are shown to be generally applicable to other human-centric video analysis tasks such as emotion recognition from user-generated videos.
\end{abstract}

\vspace{0.2cm}
\section{Introduction}
\vspace{-0.0cm}

With the recent rapid development of human-centric AI, human-centric video analysis \cite{wang2015video,wei2018deep,xu2018heterogeneous,mcduff2017large,ogawa2017human} has also begun to draw much attention from the computer vision community. Other than the conventional video content analysis that focuses on generic semantic concept analysis of video content, human-centric video analysis aims to extract, describe, and organize a wealth of information regarding the main objects of interest in most videos: humans. This topic covers a wide range of research problems such as deception detection \cite{perez2015deception,perez2015verbal}, emotion recognition from user-generated videos \cite{xu2018heterogeneous,jiang2014predicting}, personality computing \cite{wei2018deep,zhang2016bimodal}, and action recognition \cite{simonyan2014two,feichtenhofer2016convolutional,wang2016temporal,peng2016multi}. For example, it is often important to recognize the deceptive behaviors \cite{perez2015deception,perez2015verbal}, emotions \cite{xu2018heterogeneous,jiang2014predicting}, or personality traits \cite{wei2018deep,zhang2016bimodal} of the subject of a video in real-world application scenarios.

Deception detection \cite{perez2015deception,perez2015verbal} is a late addition to human-centric video analysis and still under-studied. Deception is defined as an intentional attempt to mislead others \cite{depaulo2003cues}. In our day-to-day life, deceptive behaviors occur in the form of intended lies, fabrications, omissions, misrepresentations, among others. Some deceptive behaviors are simply harmless, but others may have major threats to the society, e.g., those taking place in a courtroom. Detecting real-world human deceptive behaviors is a challenging task even for humans, and often requires well-trained human experts. A large-scale deployment of deception detection thus depends upon automated deception detection (ADD) \cite{perez2015deception,perez2015verbal}. An ADD system can find applications in many real-world scenarios including airport security screening, court trial, job interview, and personal credit risk assessment.

\begin{figure*}[ht]
\centering
\includegraphics[width=0.9\textwidth]{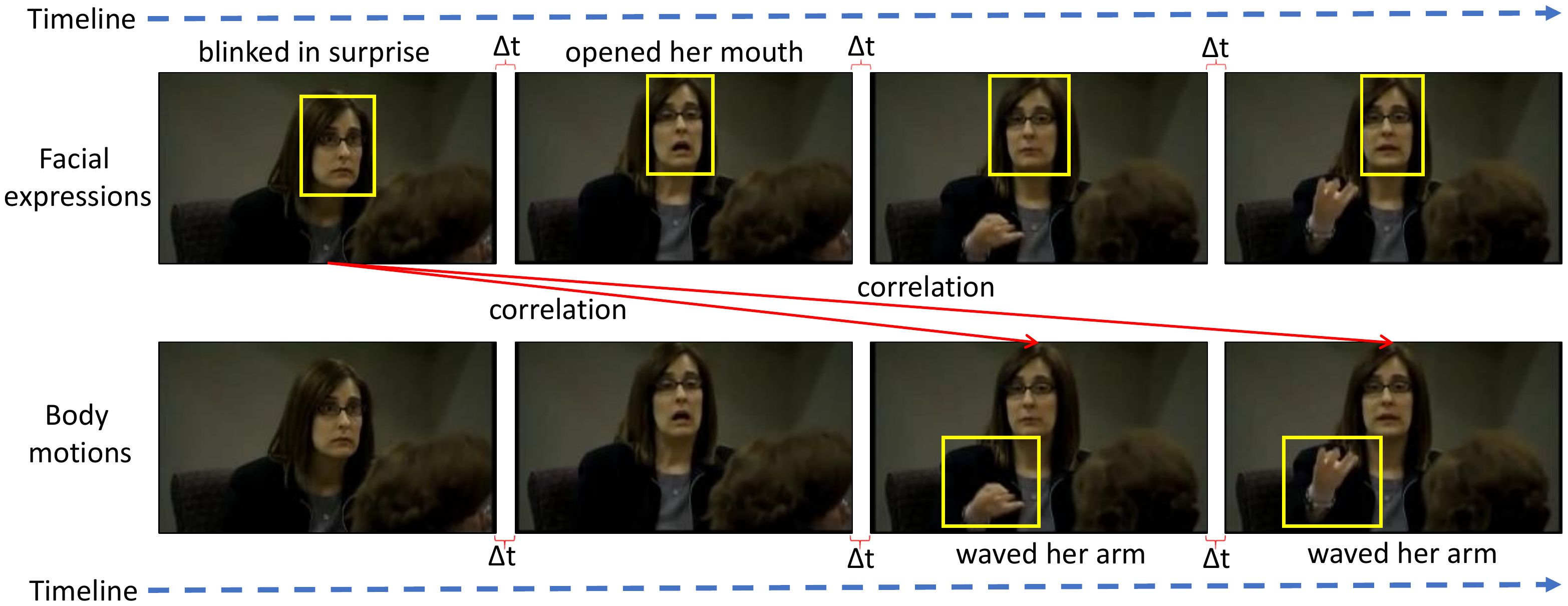}
\vspace{0.00in}
\caption{Illustration of the asynchronization between facial expressions and body motions. It can be seen that a subject who lies tends to first have a surprised expression before she/he is aroused to take body/hand movements. Notation: $\Delta \mathrm{t}=$ 2--3 frames. } \label{fig:example}
\vspace{-0.00in}
\end{figure*}

The ADD task faces two major challenges. (1) \textbf{Multi-modal fusion:}  As a subtle human behavioral trait, deception is hard to detect in real-life scenarios. Its reliable detection needs to resort to multiple  modalities including the visual, verbal, and acoustic  \cite{hirschberg2005distinguishing,howard2011acoustic,levitan2015cross,graciarena2006combining,levitan2016combining}. Among them, the visual modality is considered to be the most informative one. Multiple visual cues also exist visually. In particular, facial expressions \cite{zhang2007real,owayjan2012design} and body motions \cite{xia2007deception,michael2010motion} are typically the focus of visual analysis. An important problem thus arises: How to effectively fuse these modalities/cues? Such a fusion is not straightforward because they not only have different strengths in each individual video sequence, but also are temporally asynchronized. An example of the asynchronization between the face and body cues is shown in Figure~\ref{fig:example}.  (2) \textbf{Data scarcity:} Unlike the conventional physiological and biological methods \cite{vrij2001detecting,derksen2012control,kozel2005detecting,vrij2006detecting,gamer2014mind}, an ADD model is non-contact and non-invasive. This indirectness means that collecting large quantity of high-quality data containing samples of deceptive behaviors is critical. Earlier data collection efforts focused on human contributors in a lab or in a crowdsourcing setting. In other words, they are staged; the usefulness of these datasets for real-world deployment is thus questionable. Recently, the focus of ADD has been towards detecting deceptive behaviors from real-life data. Particularly, a new multimodal benchmark dataset of real-life videos from court trials is introduced in \cite{perez2015deception,perez2015verbal}. However, with only 121 video clips and half of them containing deception, this dataset is insufficient for training a deep neural network based model that has dominated the recent ADD approaches  \cite{gogate2017deep,krishnamurthy2018deep}.

We address both problems in this paper. To tackle the multimodal fusion problem, we propose a novel face-focused cross-stream network (FFCSN). Different from the popular two-stream networks \cite{simonyan2014two,feichtenhofer2016convolutional,wang2016temporal,peng2016multi}, our FFCSN model has two novel components: (a) Face detection is added into the spatial stream subnet to capture the facial expressions explicitly. (b) Correlation learning is performed across the spatial and temporal streams for joint deep feature learning from facial expressions and body motions. Importantly, our model is able to cope with the asynchronization/temporal inconsistency between facial expressions and body motions (see Figure~\ref{fig:example}). To address the training data scarcity problem, we introduce meta learning \cite{santoro2017simple,xie2018comparator} and adversarial learning \cite{NIPS2014_5423,li2017adversarial} into the training process of our FFCSN. Meta learning, based on the principle of learning to learn, is deployed here to improve the generalization ability of the model and avoid overfitting to the insufficient training data. In the meantime,  adversarial learning based feature synthesis is adopted as a data augmentation strategy. When these two are combined, our FFCSN can be trained effectively even with the very sparse data in the existing real-life deception detection benchmarks \cite{perez2015deception,perez2015verbal}.

Our contributions are three-fold: (1) We have proposed a novel face-focused cross-stream network (FFCSN) for joint deep feature learning from facial expressions and body motions in real-life videos. Comparing to existing two-stream networks, our FFCSN model is uniquely able to cope with the asynchronization/temporal inconsistency between facial expressions and body motions. (2) To avoid model overfitting and improve generalization ability, meta learning and adversarial learning are introduced into the training process of FFCSN. (3) We demonstrate that our FFCSN model can be easily extended to other human-centric video analysis problems such as emotion recognition from  user-generated videos \cite{xu2018heterogeneous,jiang2014predicting}. Extensive experiments are carried out on benchmark datasets and the results show that our model clearly outperforms existing state-of-the-art alternatives for both ADD and multimodal emotion recognition.

\section{Related Work}

\noindent\textbf{Video-Based Deception Detection}. Earlier works on video-based ADD are limited by the datasets which contain only staged deceptive behaviors \cite{hirschberg2005distinguishing,howard2011acoustic,levitan2015cross,graciarena2006combining,levitan2016combining}. Their usefulness for detecting real-life deception is thus in doubt. The change towards deception detection with real-life data was first advocated in \cite{fornaciari2013automatic}, where the identification of deception in statements issued by witnesses and defendants is targeted using a corpus collected from hearings in Italian courts (i.e., no visual data was available). In \cite{perez2015deception,perez2015verbal}, a new multimodal deception dataset of real-life videos from court trials was first introduced, and the combination of features extracted from different modalities is used for deception detection. Thanks to this benchmark dataset,  more advancing ADD methods \cite{jaiswal2016truth,abouelenien2017detecting,wu2018deception} have been developed to leverage multimodal features for detecting deception.

\noindent\textbf{Deep Learning for Deception Detection}. Recent ADD methods typically benefit from the latest development in deep neural networks \cite{gogate2017deep,krishnamurthy2018deep}. However, it is noted in \cite{wu2018deception} that, given the small size of the real-life ADD benchmark introduced in \cite{perez2015deception,perez2015verbal},  hand-crafted features are much better than deep features. This is not surprising: deep learning models are known to be data hungry. The real-life ADD dataset in \cite{perez2015deception,perez2015verbal} only provides around 100 video clips, which is a number of magnitudes smaller than, for example, those YouTube-collected action recognition benchmark datasets such as UCF101 \cite{soomro2012ucf101}.  Our model differs significantly from existing deep ADD models in that the data scarcity problem is addressed explicitly, based on a meta learning and adversarial learning based training strategy.  Adversarial learning \cite{NIPS2014_5423,li2017adversarial} has recently been used as a data augmentation strategy to deal with the lack of training data. However, meta-learning \cite{santoro2017simple,xie2018comparator} was originally proposed for transfer learning. Here, we re-purpose it for learning with scarce data and uniquely combine it with adversarial learning to cope with the extreme challenge of data scarcity in ADD. We show in experiments that our model outperforms \cite{jaiswal2016truth,abouelenien2017detecting,wu2018deception} by big margin, thanks to the proposed training strategy (see Tables~\ref{tab:ablation} and \ref{tab:compare}).

\noindent\textbf{Two-Stream Network}. Our FFCSN model adopts a two-stream network architecture, one for RGB still frame modeling and the other for optical flow extracted from consecutive frames. Such a two-stream architecture was originally proposed for action recognition in videos and has been popular for many human-centric video analysis tasks \cite{simonyan2014two,feichtenhofer2016convolutional}. Various improvements such as temporal segment network (TSN) \cite{wang2016temporal} and its variants \cite{Zhou_2018_ECCV,Zolfaghari_2018_ECCV} have been designed by capturing the long-range temporal structure and learning the ConvNet models with limited training samples. Similarly, \cite{peng2016multi} proposed to add faster R-CNN \cite{NIPS2015_5638} so that attention can be focused on objects detected in a video. Our FFCSN model is different from existing two-stream models in that: (1) face detection is added into the spatial stream subnet to capture the facial expressions explicitly; (2) correlation learning is performed across the spatial and temporal streams to cope with the temporal inconsistency between facial expressions and body motions for ADD.

\noindent\textbf{Video-Based Emotion Recognition}. Deception detection is closely related to emotion recognition: deception could be considered as a specific emotion state of humans, albeit it is much more subtle and harder to detect than others such as happy and angry. Although emotion recognition with still face images has been well studied in previous works, emotion recognition from user-generated videos \cite{jiang2014predicting} is still a challenging problem. In particular, because of the complicated and unstructured nature of user-generated videos and the sparsity of video frames that express the emotion content, it is often hard to understand emotions conveyed in user-generated videos. To address this challenging problem, multi-modal fusion and knowledge transfer approaches have been proposed in recent works \cite{pang2015deep,xu2016video,zhang2018recognition,xu2018heterogeneous}. In this paper, we show that our FFCSN model can be easily extended to emotion recognition from user-generated videos, with state-of-the-art results achieved.

\begin{figure*}[t]
\centering
\includegraphics[width=0.96\textwidth]{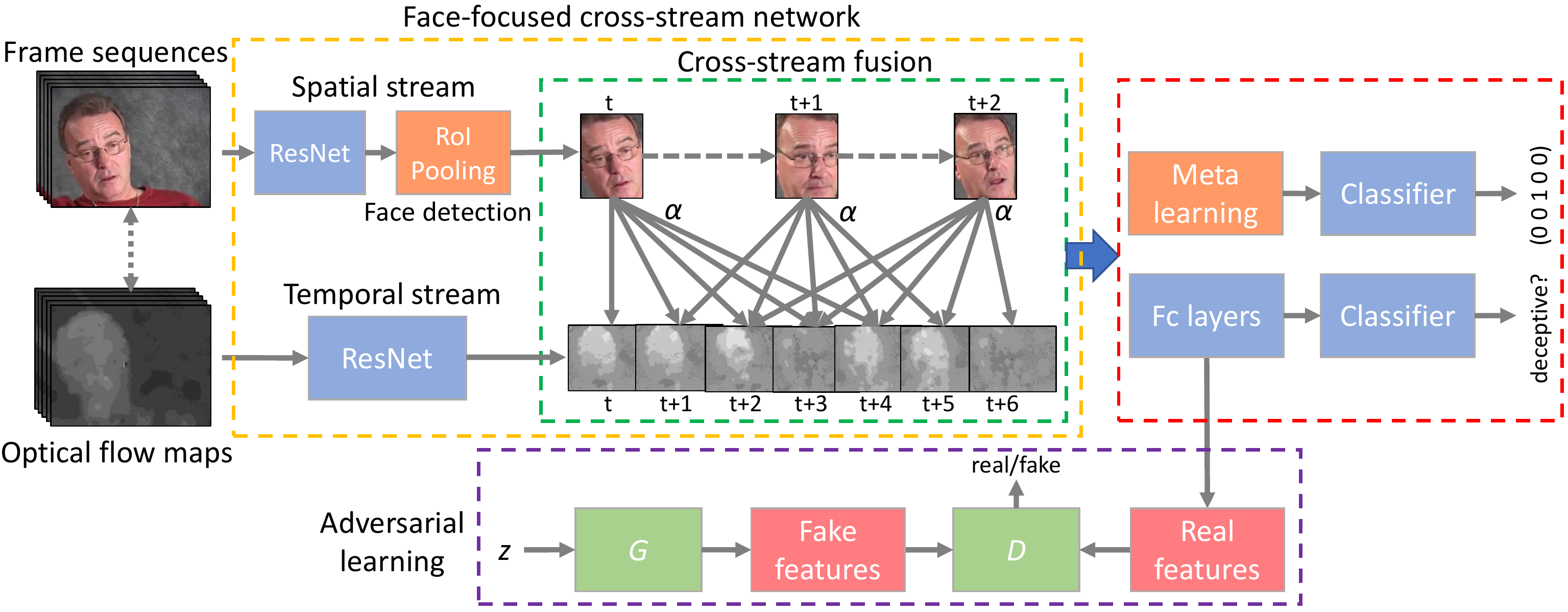}
\vspace{0.05in}
\caption{An overview of our full FFCSN model for video-based ADD. Three main modules are included in our full FFCSN model: face-focused cross-stream base network, meta learning module, and adversarial learning module. }\label{fig:pipeline}
\vspace{-0.05in}
\end{figure*}

\section{Methodology}

As illustrated in Figure~\ref{fig:pipeline}, our full FFCSN model for video-based ADD consists of three main modules: face-focused cross-stream network including a facial expression branch as well as a body motion branch, meta learning module, and adversarial learning module. In the following, we give the details of the three main modules.

\subsection{Cross-Stream Network Module}

In this work, we focus on joint deep feature learning from facial expressions and body motions for video-based ADD. Different from the traditional video-based action recognition, the facial expressions and body motions of a subject are found to be related to his/her deceptive behaviors \cite{zhang2007real,owayjan2012design,xia2007deception,michael2010motion}, rather than the whole frame appearance. Therefore, we choose to modify the original two-stream temporal segment network \cite{wang2016temporal} designed for video-based action recognition by replacing its appearance branch with a face expression branch (see Figure~\ref{fig:pipeline}).

\subsubsection{Cross-Stream Base Network}

The spatial stream (i.e. face expression branch) is a face detection model based on the popular faster R-CNN \cite{NIPS2015_5638}. This branch follows the deep learning framework of faster R-CNN, which has been shown to achieve state-of-the-art results in generic object detection. As illustrated in Figure~\ref{fig:pipeline}, it essentially consists of two parts: (1) a region proposal network (RPN) for generating a list of region proposals which may contain objects, called regions of interest (RoIs); (2) a R-CNN network for classifying the regions of each frame into objects and refining the boundaries of these regions. The two parts share common parameters in the convolutional layers used for feature extraction, allowing it to accomplish the face detection task efficiently.

In our model, faster R-CNN is generalized for both face detection and expression feature extraction. Note that the traditional faster R-CNN takes only 9 anchors, which sometimes fails to recall small objects. For our face detection task, however, small faces tend to be fairly common. We thus add a size group of 64$\times$64 and increase the number of anchors to 12. In this paper, the RPN batchsize is set to 256, and the ResNet50 \cite{he2016cvpr} is used as the backbone model for the face expression branch.

The temporal stream (i.e. body motion branch) operates on a stack of consecutive warped optical flow fields to capture the motion information. Inspired by the representative work on improved dense trajectories \cite{Wang2013iccv}, we extract the warped optical flow by first estimating the homography matrix and then compensating the camera motion. This branch can thus avoid concentrating on the camera motion but not on the body motion. As shown in Figure~\ref{fig:pipeline}, ResNet50 is used to compute the temporal feature maps.

\begin{figure}[t]
\centering
\includegraphics[width=0.96\columnwidth]{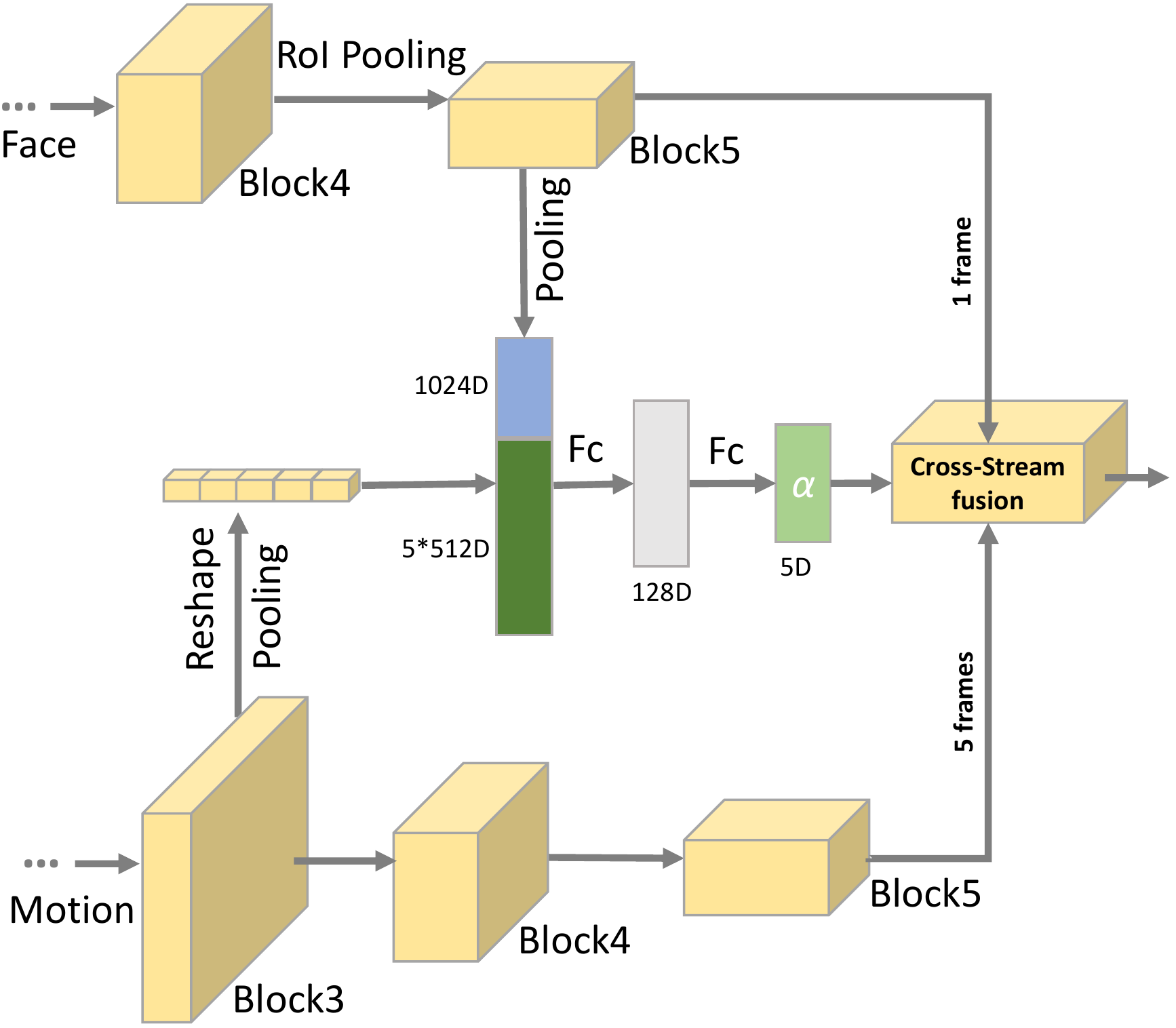}
\caption{Architecture of our cross-stream fusion block. }\label{fig:ffcsn}
\vspace{-0.05in}
\end{figure}

\vspace{-0.1cm}
\subsubsection{Cross-Stream Fusion}
\vspace{-0.1cm}

In our cross-stream base network, one stream focuses on the face that is only a part of the whole frame, and the other focuses on the body motion that is captured using multiple whole frames. These two parts are clearly complementary to each other. We thus combine the two branches by cross-stream fusion as illustrated in Figure~\ref{fig:ffcsn}. Importantly, the fusion is based on deep correlation analysis rather than simply concatenating the feature vectors extracted from the two streams as in conventional two-stream networks. Specifically, to cope with the asynchronization/temporal inconsistency between facial expressions and body motions (see Figure~\ref{fig:example}), we choose to learn the correlation among adjacent frames (only 5 adjacent frames are considered here). For the spatial stream, we downsample the feature maps of the final residual block of ResNet50 \cite{he2016cvpr} in the dimension of depth and obtain a 1024-dimension feature vector. For the temporal stream, given that five motion frames are matched to one face frame, we utilize the reshape pooling to obtain one 5$\times$512-dimension feature vector after the third residual block of ResNet50. The outputs of the two streams are then concatenated and fed into two fully-connected layers (with the dimension of 128 and 5, respectively). Finally, we compute the correlation scores $\alpha=[\alpha_1,...,\alpha_5]$ for the five face-motion pairs using the softmax function, and weight them with $\alpha$ for final two-stream fusion.

To extract deep visual features from a long-term video, our model essentially works on a sequence of short snippets sparsely sampled from the entire video. After each snippet of this sequence predicts its own result, a consensus among all snippets is obtained as the final video-level prediction. For all obtained video-level predictions, we can define a segmental consensus classification loss similar to that of temporal segment network \cite{wang2016temporal}. Specifically, we divide each video into $K$ segments $\{S_1, S_2, \ldots, S_K\}$ of equal duration. From each segment $S_k$ ($k=1,...,K$), we then randomly sample a short snippet $T_k$. In our problem, the short snippet $T_k$ consists of one spatial frame (denoted as $T_k(\mathrm{sf}_1)$) and five temporal frames (denoted as $T_k(\mathrm{tf}_1),...,T_k(\mathrm{tf}_5)$). Suppose that all snippets/frames have been represented as feature maps here. We thus have $T_k=[T_k(\mathrm{sf}_1), \sum_{j=1}^5 \alpha_j T_k(\mathrm{tf}_j)]$. Let $\mathcal{F}(T_k; W)$ denote the classification probability predicted by our model with parameters $W$ for $T_k$. The outputs of all short snippets are combined by the segmental consensus function $\mathcal{E}$ to obtain a consensus of prediction among them. With the softmax loss, the overall loss of our model is defined as:
\begin{small}
\begin{align}
L_{BASE}(y, E) = -\sum_{i=1}^{N_c} y_i(E_i - \log\sum_{j=1}^{N_c} \exp E_j),
\label{eq:lossbase}
\end{align}
\end{small}
\hspace{-4.5pt}where $E$ is the segment consensus computed by $E = \mathcal{E}(\mathcal{F}(T_1; W), \mathcal{F}(T_2; W), \ldots, \mathcal{F}(T_K; W))$, ${N_c}$ is the number of target classes (${N_c}=2$ in our problem), and $y_i$ is the ground truth label with respect to class $i$. We define the consensus function $\mathcal{E}$ with average pooling, as in \cite{wang2016temporal}.

\vspace{-0.0cm}
\subsection{Meta Learning Module}
\vspace{-0.0cm}

\begin{figure}[t]
\centering
\includegraphics[width=0.99\columnwidth]{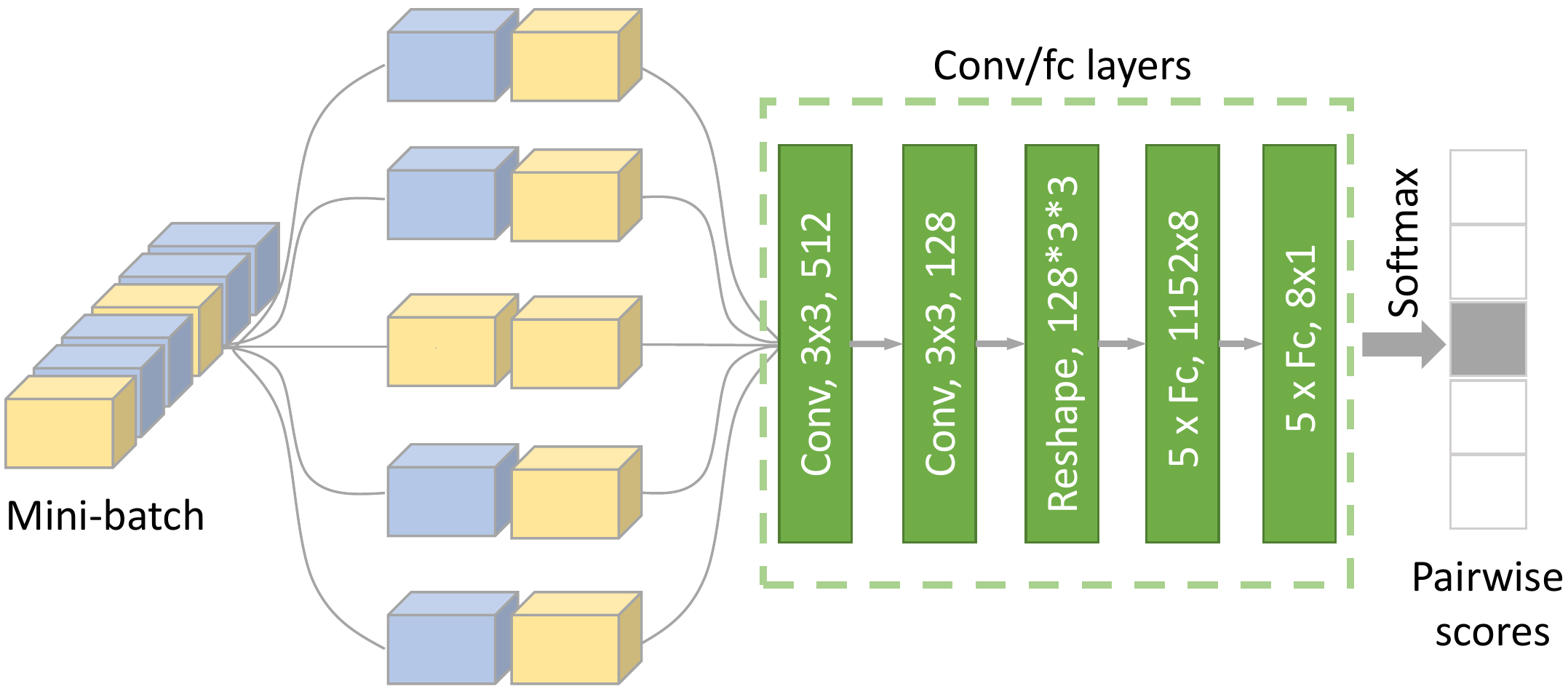}
\caption{Architecture of the meta learning module used in our full FFCSN model for video-based ADD. See text for details. }\label{fig:relation}
\vspace{-0.0in}
\end{figure}

Deception detection is a challenging task due to the subtle differences between truthful and deceptive behaviors. Learning to differentiate the two types of behaviors with only a handful of samples of each is extremely challenging. This is especially true when the behaviors are modeled with deep neural networks with a large number of model parameters. To deal with the data scarcity problem, we propose to use meta learning \cite{santoro2017simple,xie2018comparator} to train our FFCSN (see Figure~\ref{fig:pipeline}). To best utilize the limited training samples, we introduce pair-wise comparison of them. Specifically, our cross-stream base network can be viewed as the encoding submodule $f$ of our meta learning module. A comparison submodule $g$ is then introduced for meta learning. The meta learning pipeline is illustrated in Figure~\ref{fig:relation}. Examples of the two classes (yellow deceptive and blue truthful) are shown in different colors. In this case, the meta-train set contains five samples (four truthful and one deceptive). The deceptive sample in the meta-validation set is used to form five pairs with the meta-train samples. The final model output, after softmax, is a 5D logit vector supervised to produce a close-to-one value in the third element and close-to-zero values in all other elements. This meta-learning pipeline turns a two-class (deceptive/deceptive) classification problem into a multi-case classification problem and makes full use of the limited training samples.

Formally, in each mini-batch (with the mini-batch size $N_b$), videos $x_i ~(i=1, 2, \ldots, N_b)$ are fed through the encoding submodule $f$, which outputs the concatenated feature maps $f(x_i)~(i=1, 2, \ldots, N_b)$. We split the videos in the mini-batch into the meta-train and meta-validation sets. A sample $x_a$ is randomly chosen from the meta-validation set. The output $f(x_a)$ is combined with each $f(x_j)~(j \not=a)$ in the meta-train set using the operator $\mathcal{C}(f(x_a), f(x_j))$. In our meta learning module, we set $\mathcal{C}(\cdot,\cdot)$ as the concatenation of feature maps in the dimension of depth. The combined feature maps of the sample pairs are fed into the comparison submodule $g$, which produces a pairwise score representing the similarity between $x_a$ and $x_j$. We thus generate the pairwise scores for each mini-batch as:
\begin{small}
\begin{align}
r_{a,j} = g(\mathcal{C}(f(x_a), f(x_j))), j \not= a.
\label{eq:relationscore}
\end{align}
\end{small}

We train our meta learning module by fitting the pairwise score $r_{a,j}$ to the ground truth pairwise similarity with a cross entropy loss as follows:
\begin{small}
\begin{align}
L_{ML} \hspace{-0.03in}=\hspace{-0.03in} \frac{-1}{N_b - 1}\sum_{j\not=a} y_j\log(r_{a,j}) \hspace{-0.03in}+\hspace{-0.03in} (1 - y_j)\log(1 - r_{a, j}),
\label{eq:lossrn}
\end{align}
\end{small}
\hspace{-3pt}where $y_j = 1$ if $(x_a, x_j)$ is an intra-class sample pair, and $y_j = 0$ if $(x_a, x_j)$ is an inter-class sample pair.

As illustrated in Figure~\ref{fig:relation}, the encoding submodule of our meta learning module is just our cross-stream base network. In the following, we give the details of the comparison submodule of our meta learning module. Specifically, the comparison submodule consists of two convolutional blocks and two fully-connected layers: (1) Each convolutional block has a $3 \times 3$ convolution layer followed by batch normalization and RELU activation. The number of filters of the convolution layer in the first block is 512 and the number of filters in the second block is 128. The output size of the two blocks is $128 \times 3 \times 3 = 1,152$. (2) The two fully-connected layers are 8 and 2 dimensional, respectively.

\subsection{Adversarial Learning Module}

In this paper, we aim to synthesize feature vectors for data augmentation in the ADD task. Note that synthesizing raw videos explicitly is an unsolved problem in itself. Therefore, we choose to generate a 256-dimension feature vector for each synthesized video instead, which is a much easier task. In particular, we propose to synthesize \emph{fake feature vectors and attack the classifier} for deception detection during training of our full FFCSN model, in order to overcome the training data scarcity problem.

Adversarial training involves a discriminator and a generator. In our case, the discriminator network aims to classify the inputs into two classes: real or fake. In this paper, the observed variable $x$ is the 256-dimensional vector produced by our cross-stream base network. Given that the discriminator network $D$ consists of 3 fully-connected layers with the ELU activation, $D(x)$ thus denotes the probability that $x$ comes from the real (but not fake) class.

As for the generator network $G$ of our adversarial learning module, the input 32-dimensional noise $z$ is sampled from a zero-mean Gaussian distribution $p_z(z)$ with the standard deviation 1. We use 3 hidden layers to represent $G$ with the size 32, 64, and 256, respectively. The first fully-connected layer uses the ELU activation, and the second fully-connected layer uses the sigmoid activation. $G(z)$ denotes a generated sample drawn from the data space.

The adversarial training of $D$ and $G$ can be formulated as the following min-max problem:
\begin{small}
\begin{align}
\min_G \max_D &~L_{AL}(G,D)=\mathbb{E}_{x\sim p_{\mathrm{data}}(x)}[\log D(x)]\nonumber\\
&+ \mathbb{E}_{z\sim p_z(z)}[\log(1-D(G(z)))], \label{eq:lossgan}
\end{align}
\end{small}
\hspace{-3pt}where $L_{AL}$ denotes the loss function of our adversarial learning module and $p_{\mathrm{data}}(x)$ denotes the data distribution.

\subsection{Training Process}

Our full FFCSN model for video-based ADD is trained using an end-to-end training strategy. The loss function of our full FFCSN model is defined as follows:
\begin{small}
\begin{align}
L = L_{BASE}+\beta_1 L_{ML}+ \beta_2 L_{AL},
\end{align}
\end{small}
\hspace{-3pt}where $\beta_1$ and $\beta_2$ denote the hyper-parameters. In this paper, we empirically set $\beta_1=\beta_2=1$ in all experiments.

\section{Experiments}

\subsection{Video-Based Deception Detection}
\label{sect:exp:add}

\subsubsection{Dataset and Setting}

\noindent\textbf{Real-Life Dataset}. We evaluate our full FFCSN model for deception detection on a real-life multimodal dataset \cite{perez2015deception}. This dataset consists of 121 court room trial video clips. Since videos from this trial dataset are collected under unconstrained conditions, we need to cope with the change of the viewing angle of the person, the variation in video quality, and the background noise. In this paper, we select a subset of 104 videos from the original trial dataset, including 50 truthful videos and 54 deceptive videos, as in \cite{wu2018deception}.

\noindent\textbf{Evaluation Setting}. Our dataset consists of only 58 identities. Since the number of identities is smaller than the number of video clips, the same identity often appear in both  deceptive  truthful clips. When the videos of the same identity are divided into both the training and test sets, a deception detection method tends to suffer from over-fitting to identities. To address this over-fitting issue, we perform 10-fold cross validation over identities (but not over video samples) as in \cite{wu2018deception}, which ensures that the identities in the test set have no overlap with that in the training set.

\noindent\textbf{Evaluation Metrics}. To evaluate the performance of a deception detection method, we compute two metrics as follows: (1) ACC -- the classification accuracy (ACC) over the test video samples; (2) AUC -- the area under the precision-recall curve (AUC) over the test set, which is originally defined to cope with the imbalance of the positive and negative classes. The former has been widely used in previous research on deception detection \cite{perez2015deception,perez2015verbal,jaiswal2016truth,gogate2017deep}, while the latter is mainly used in recent works \cite{wu2018deception,krishnamurthy2018deep}.

\noindent\textbf{Network Initialization}. We pretrain the face branch of our cross-stream base network using the WIDER-FACE \cite{yang2016wider} and CK+ \cite{LC10} datasets, and then pretrain the motion branch of our cross-stream base network as in \cite{wang2016temporal} on the UCF101 \cite{soomro2012ucf101} dataset. Moreover, for $G$ and $D$ of the adversarial learning module, we adopt the Kaiming initialization. All the other layers are randomly initialized by drawing weights from a zero-mean Gaussian distribution with the standard deviation 0.01 (along with zero biases).

\noindent\textbf{Implementation Details}. After network initialization, our full FFCSN model is trained in an end-to-end manner using back-propagation and stochastic gradient descent. The learning rate is set to 0.0005 for the first 10 epochs, and then reduced to one tenth with a step size of 10 (epochs). The maximum number of epochs is set to 100. A momentum of 0.9 and a weight decay of 0.01 are also set for model training. We train our full FFCSN model on two Tesla K40 GPUs, with the batch size 12. Our implementation is developed within the PyTorch framework.

\begin{table}[t]
\vspace{0.05in}
\begin{center}
\begin{small}
\tabcolsep0.5cm
\begin{tabular}{l|c|c}
\hline
Model & ACC & AUC\\
\hline
Face                   &  84.33 & 84.11 \\
Motion                 &  86.00 & 88.63 \\
Face+Motion            &  88.21 & 90.57 \\
Face+Motion+CL         &  89.16 & 91.89 \\
Face+Motion+CL+ML      &  92.33 & 95.83 \\
Face+Motion+CL+ML+AL   & \textbf{93.16} & \textbf{96.71} \\
\hline
\end{tabular}
\end{small}
\end{center}
\vspace{-0.05in}
\caption{Ablation study results (\%) for our full FFCSN model. }
\label{tab:ablation}
\vspace{-0.1in}
\end{table}

\vspace{-0.2cm}
\subsubsection{Ablation Study Results}
\vspace{-0.1cm}

To show the contribution of each main module of our full FFCSN model, we make comparison to its five simplified versions: (1) Face - Only the face branch of our cross-stream base network used for ADD; (2) Motion - Only the motion branch of our cross-stream base network used for ADD; (3) Face+Motion -- our cross-stream base network including the face and motion branches (but without cross-stream correlation learning); (4) Face+Motion+CL -- our cross-stream base network with cross-stream correlation learning (CL); (5) Face+Motion+CL+ML -- our cross-stream base network further boosted with meta learning (ML). Our full model including adversarial learning (AL) is denoted as Face+Motion+CL+ML+AL.

\begin{figure}[t]
\centering
\includegraphics[width=0.90\columnwidth]{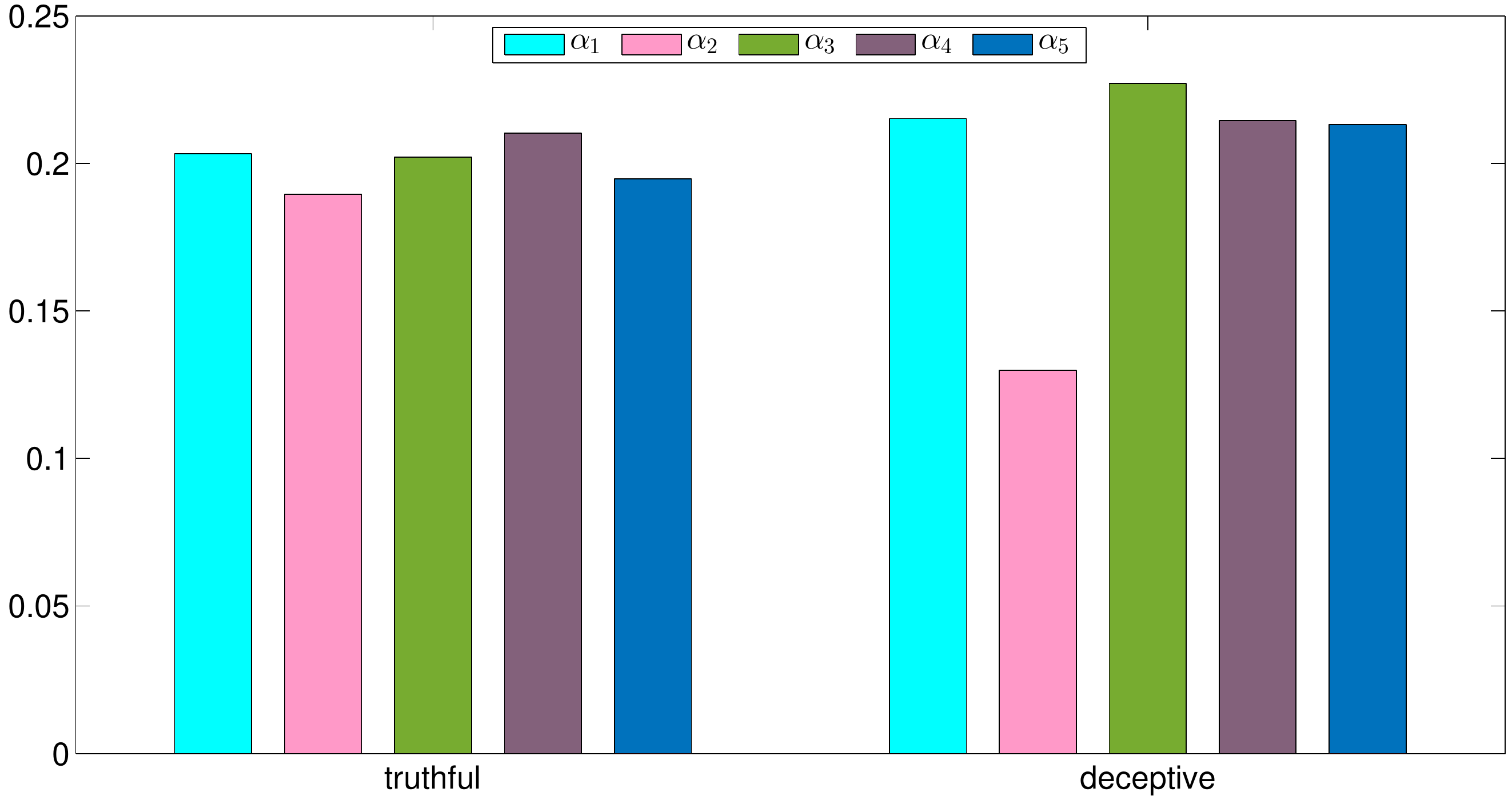}
\caption{Illustration of the two mean correlation distributions (i.e. $\alpha=[\alpha_1,...,\alpha_5]$) obtained with our cross-stream network averaged over the truthful and deceptive test samples, respectively. Note that the deceptive test samples with $\alpha_1,\alpha_2<0.15$ occupy 14.2\% of all deceptive test samples, directly showing the \emph{temporal inconsistency} of deceptive behaviors.  }\label{fig:corre}
\vspace{-0.05in}
\end{figure}

The ablation study results are presented in Table~\ref{tab:ablation}. It can be seen that: (1) The performance continuously increases when more modules are used for ADD, showing the contribution of each module. (2) The improvements achieved by Face+Motion over Face/Motion show that both face expression and body motion are important cues for ADD. (3) Both ML and AL clearly lead to performance improvements, which provides evidence that they have a good ability of alleviating the training data scarcity. (4) The effectiveness of cross-stream correlation learning is validated by the comparison Face+Motion+CL vs. Face+Motion. This is further supported by Figure~\ref{fig:corre}, where our cross-stream correlation learning is found to learn quite different correlation distributions for the truthful/deceptive classes. That is, the learned correlations indeed improve the discriminativeness of deep visual features for deception detection.

\begin{table}[t]
\vspace{0.05in}
\begin{center}
\begin{small}
\tabcolsep0.35cm
\begin{tabular}{l|c|c}
\hline
Model  & ACC & AUC\\
\hline
\cite{perez2015deception} (visual+verbal)   & 75.20 & -- \\
\cite{perez2015verbal} (visual+verbal)    & 77.11 & -- \\
\cite{jaiswal2016truth} (visual+acoustic+verbal) & 78.95 & --\\
\cite{gogate2017deep} (visual+acoustic+verbal)   & 96.42 & -- \\
\cite{wu2018deception} (visual+acoustic+verbal)    & -- & 92.21 \\
\cite{krishnamurthy2018deep} (visual+acoustic+verbal)  & 96.14 & 97.99 \\
\hline
Ours (visual)       & 93.16    &  96.71 \\
Ours (visual+acoustic+verbal) & \bf97.00 & \bf99.78 \\
\hline
\end{tabular}
\end{small}
\end{center}
\vspace{-0.05in}
\caption{Comparative results (\%) for video-based ADD. Note that extra \emph{human annotated} micro-expressions are used in \cite{wu2018deception,krishnamurthy2018deep}. }
\label{tab:compare}
\vspace{-0.1in}
\end{table}

\vspace{-0.2cm}
\subsubsection{Comparative Results}
\vspace{-0.1cm}

We further make comparison to the state-of-the-art alternatives \cite{jaiswal2016truth,gogate2017deep,wu2018deception,krishnamurthy2018deep}. Since all of these methods are multimodal, we also include the acoustic and verbal modalities:

\noindent\textbf{Acoustic Feature Learning}. We extract the spectrum map from each wave audio of 44,100 Hz sampling rate, and convert each spectrum map into images of fixed size using a sliding window with the window size 300. By taking only the last 300 dimensions along the spectrum height, we obtain a set of samples of the size 300*300. These samples are finally used to train ResNet50. For robust training, ML and AL are similarly exploited for acoustic feature learning.

\noindent\textbf{Verbal Feature Learning}. We segment the transcript of each video to words, and then employ the word2vec technique \cite{goldberg2014word2vec} to convert each word into a 300-dimensional feature vector. The feature vectors of all words are averaged as the verbal feature vector of a video. The average vector is fed into three layers of fully connected layers (of the size 300*128, 128*64, and 64*32), resulting in a final vector of 32 dimensions. For robust verbal feature learning, ML and AL are also used like visual feature learning.

\begin{figure}[t]
\centering
\includegraphics[width=0.94\columnwidth]{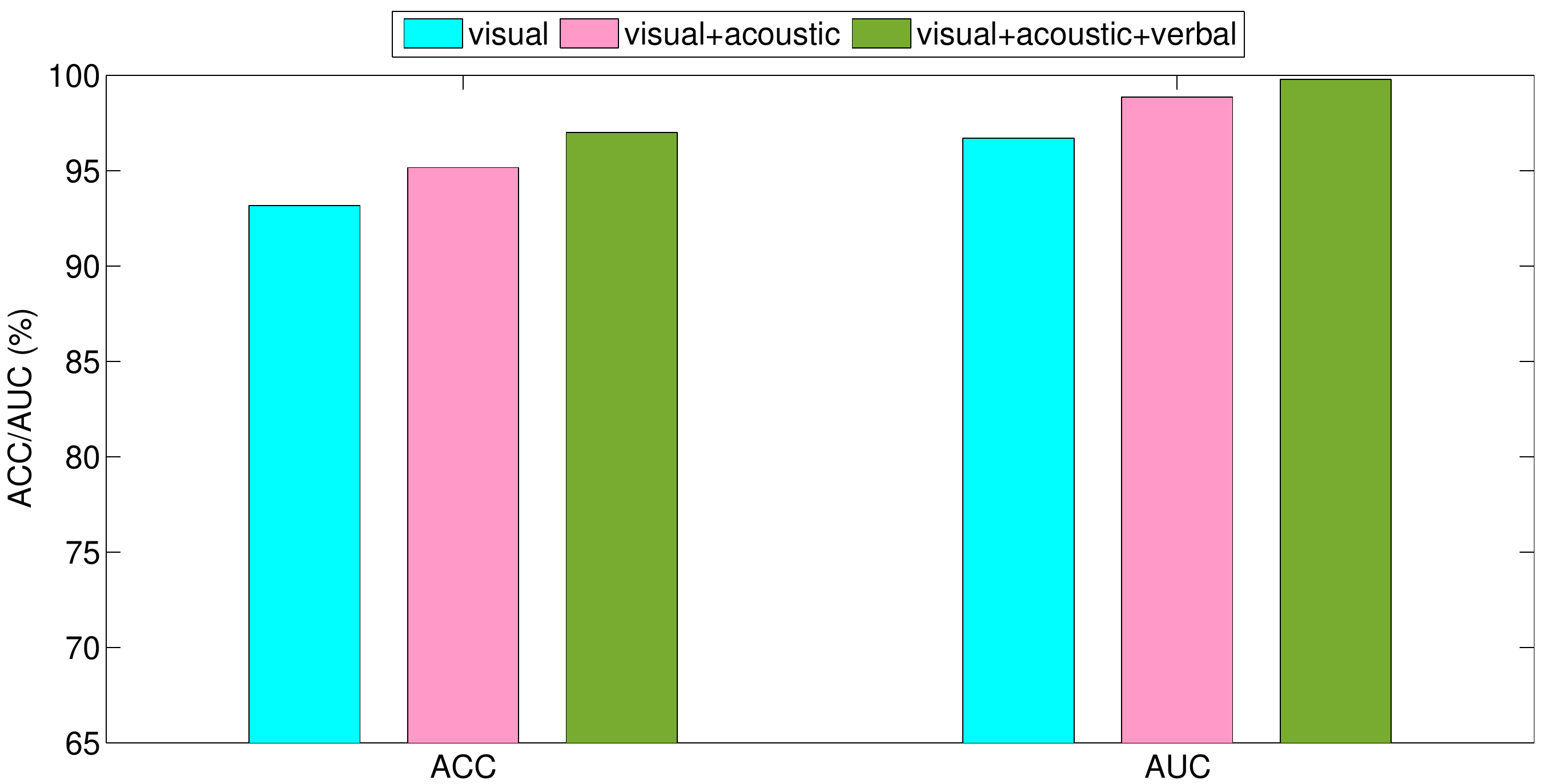}
\caption{Comparative results obtained by multi-modality fusion. }\label{fig:fusion}
\vspace{-0.00in}
\end{figure}

The comparative results on the real-life benchmark dataset \cite{perez2015deception} are given in Table~\ref{tab:compare}. We observe that: (1) Our robust deep feature learning approach clearly performs the best under the multimodal setting, validating the effectiveness of exploiting ML and AL for addressing the training data scarcity issue associated with real-life ADD. (2) When only the visual modality is concerned, our robust deep feature learning approach even outperforms the state-of-the-art multimodal deception detection method \cite{wu2018deception}. (3) Our multimodal approach achieves performance improvements over the latest deep learning methods \cite{gogate2017deep,krishnamurthy2018deep}, due to the extra use of ML and AL in our approach. In addition, we also provide the comparative results of modality fusion for our approach in Figure~\ref{fig:fusion}. As expected, our approach is shown to obtain more significant improvements when more modalities are used for deception detection in videos.

\begin{figure}[t]
\vspace{0.05in}
\centering
\includegraphics[width=0.94\columnwidth]{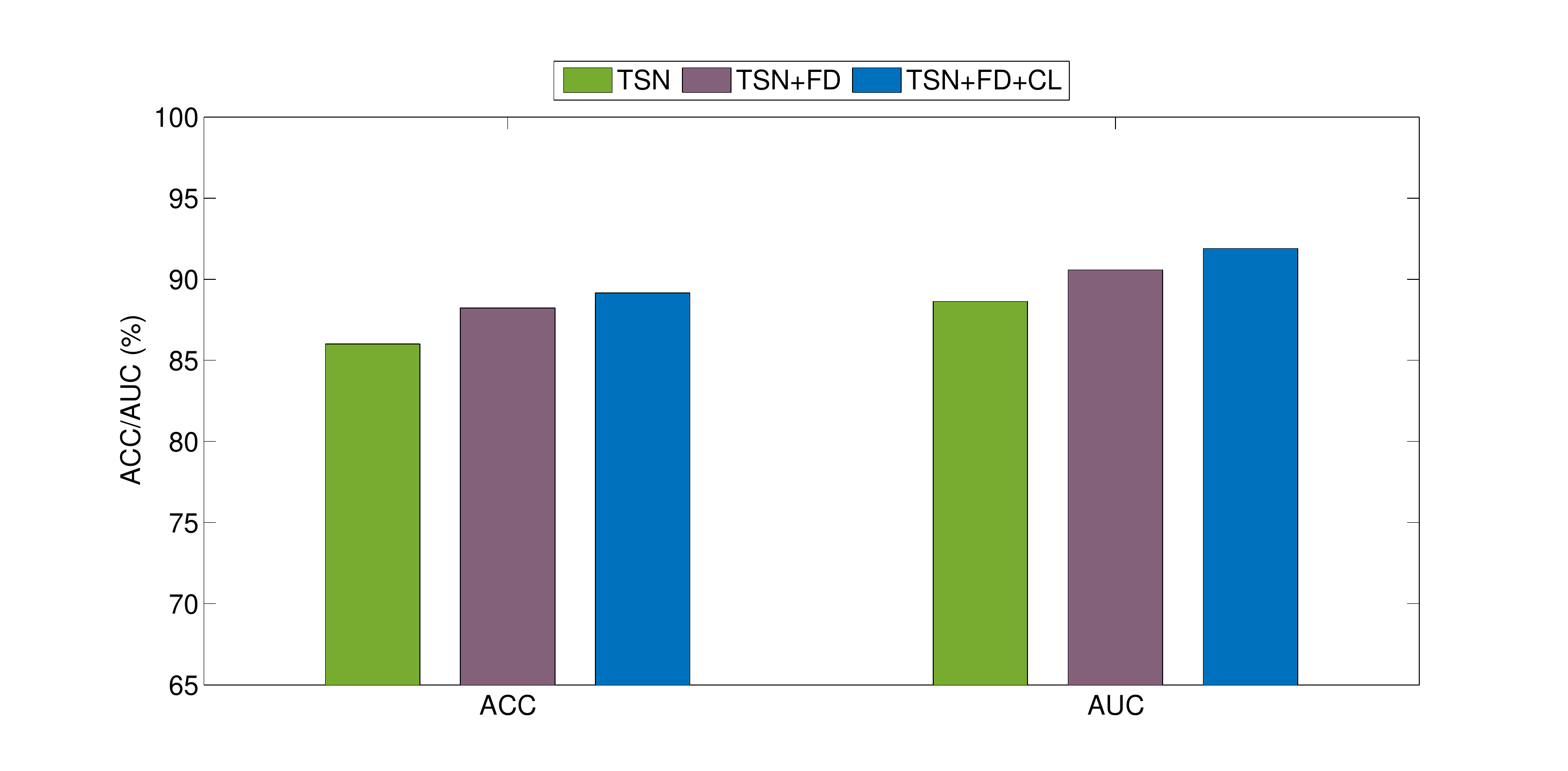}
\caption{Comparison to temporal segment network (TSN). }\label{fig:tsn}
\vspace{-0.05in}
\end{figure}

\vspace{-0.1cm}
\subsubsection{Further Evaluations}
\vspace{-0.1cm}

\noindent\textbf{Comparison to Temporal Segment Network}. Different from the state-of-the-art temporal segment network (TSN) \cite{wang2016temporal}, our FFCSN model has two novel components: face detection and correlation learning. To show the contribution of these two components, we obtain two variants of our FFCSN model by adding face detection (FD) and correlation learning (CL) into TSN: (1) TSN+FD: face detection is added to the spatial stream of TSN; (2) TSN+FD+CL: cross-stream correlation learning is further used to boost TSN+FD. The comparative results in Figure~\ref{fig:tsn} clearly show that both components are effective for ADD.

\begin{figure}[t]
\centering
\includegraphics[width=0.88\columnwidth]{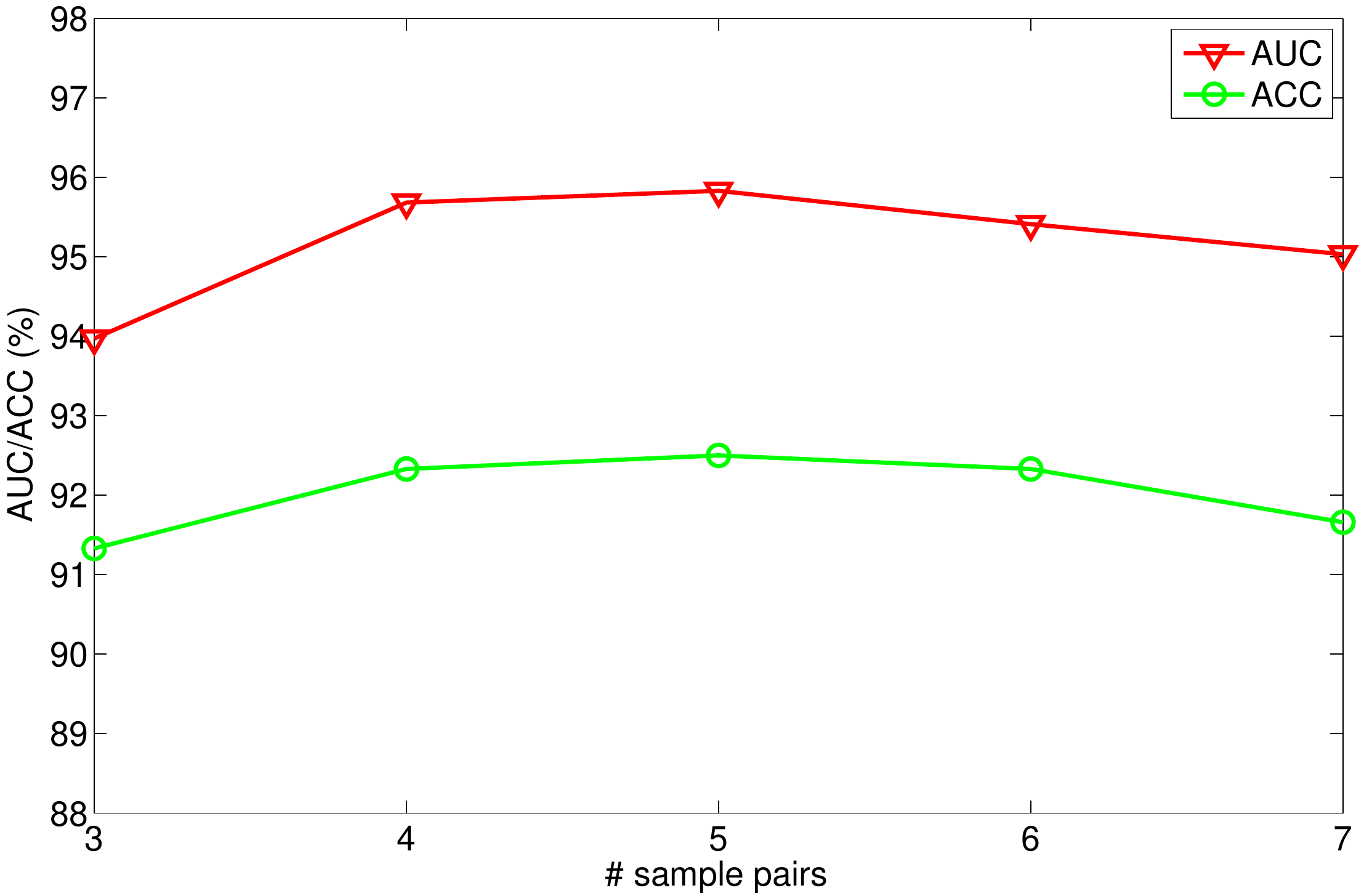}
\caption{Illustration of the effect of the number of sample pairs used for pairwise comparison on the performance of meta learning. }\label{fig:relationk}
\vspace{-0.05in}
\end{figure}

\begin{figure}[t]
\centering
\includegraphics[width=0.90\columnwidth]{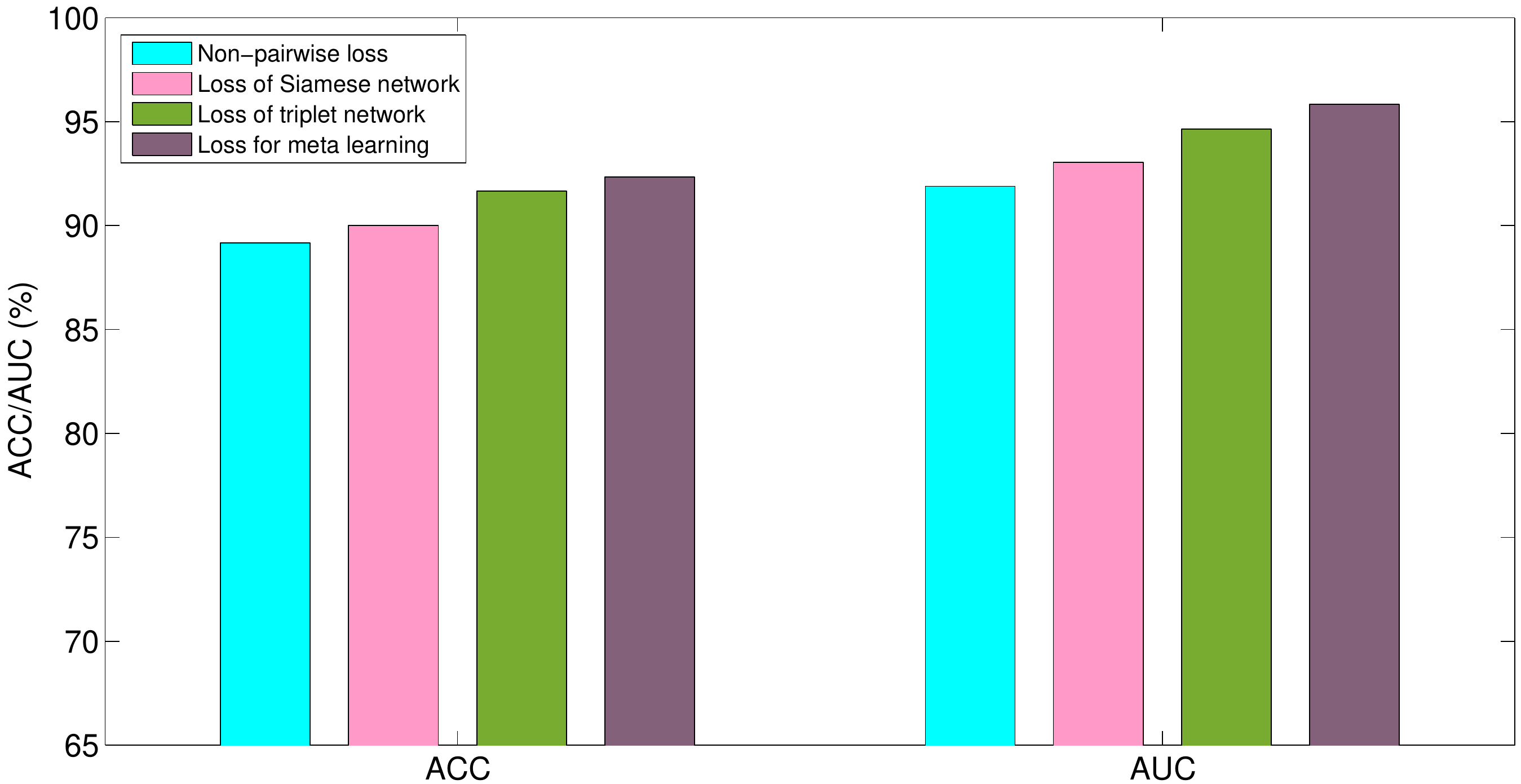}
\caption{Comparative results obtained by employing different losses for pairwise comparison. } \label{fig:pairwiseloss}
\vspace{-0.1in}
\end{figure}

\noindent\textbf{Model Selection for Meta Learning}. As illustrated in Figure~\ref{fig:relation}, the number of sample pairs in each sampled task in the meta-learning pipeline  is empirically set to 5. To evaluate the impact of the task size on the model performance,   Figure~\ref{fig:relationk} compares different task sizes. It can be clearly seen that our model approaches the peak at 5, but it is in general insensitive to the task size selection.

\noindent\textbf{Alternative Losses for Pairwise Comparison}. In this paper, our loss defined in Eq.~(\ref{eq:lossrn}) is used for pairwise comparison. To show the effectiveness of such loss, we compare it to two typical pairwise losses under the same setting: loss of Siamese network \cite{chopra2005learning}, and loss of triplet network \cite{hoffer2015deep}. The conventional non-pairwise loss is also included as the baseline. The comparative results in Figure~\ref{fig:pairwiseloss} show that: (1) All three pairwise losses clearly lead to better results than the conventional non-pairwise loss, validating the effectiveness of pairwise comparison for deception detection. (2) The loss defined in Eq.~(\ref{eq:lossrn}) performs the best among the three pairwise losses, i.e., the meta learning module is more capable of modelling the complicated relationships among video samples than the Siamese network and triplet network.

\begin{table}[t]
\vspace{0.05in}
\begin{center}
\begin{small}
\tabcolsep0.35cm
\begin{tabular}{c|l|c}
\hline
Model       &  multimodal  & ACC\\
\hline
\cite{jiang2014predicting}  & visual+acoustic+attribute   &  46.1 \\
\cite{pang2015deep}         & visual+acoustic+attribute   &  51.1 \\
\cite{zhang2018recognition} & visual+attribute            &  52.5 \\
\cite{xu2018heterogeneous}  & visual+acoustic             &  52.6 \\
\cite{xu2016video}          & visual+acoustic             &  52.6 \\
\hline
Ours                        & visual                      &  \bf57.8 \\
\hline
\end{tabular}
\end{small}
\end{center}
\vspace{-0.05in}
\caption{Comparative results (\%) of video-based emotion recognition on the YouTube-8 dataset.  }
\label{tab:emotion}
\vspace{-0.0in}
\end{table}

\vspace{-0.1cm}
\subsection{Video-Based Emotion Recognition}
\vspace{-0.0cm}

\vspace{-0.0cm}
\subsubsection{Dataset and Setting}
\vspace{-0.1cm}

The YouTube-8 dataset \cite{jiang2014predicting} is used for performance evaluation. This dataset consists of 1,101 videos (downloaded from YouTube) annotated with 8 basic emotions: anger, anticipation, disgust, fear, joy, sadness, surprise, and trust. We randomly generate 10 train/test splits, each using 2/3 of the dataset for training and 1/3 for testing. The averaged recognition accuracy (ACC) over 10 random train/test splits is used as the evaluation metric. Our FFCSN model is trained exactly the same as in Section~\ref{sect:exp:add}, and only visual features are extracted from raw videos for emotion recognition.

\vspace{-0.1cm}
\subsubsection{Comparative Results}
\vspace{-0.0cm}

We compare our FFCSN model to the state-of-the-art alternatives \cite{jiang2014predicting,pang2015deep,xu2016video,zhang2018recognition,xu2018heterogeneous}. The comparative results are presented in Table~\ref{tab:emotion}. We have the following observations: (1) Our FFCSN model achieves significant improvements over the state-of-the-art models, validating the effectiveness of our face-focused cross-stream network for emotion recognition from user-generated videos. Note that the biggest challenge of this emotion recognition task lies in the complicated and unstructured nature of user-generated videos and the sparsity of video frames that express the emotion content. Our FFCSN model is clearly effective in overcoming this challenge. (2) The improvements obtained by our FFCSN model are really impressive, given that only visual features are extracted by our model, whilst at least two modalities are used by all other models.

\vspace{-0.0cm}
\section{Conclusion}
\vspace{-0.0cm}

In this paper, we have investigated the challenging problem of deception detection from real-life videos. For joint deep feature learning from facial expressions and body motions, we have proposed a novel face-focused cross-stream network (FFCSN). Importantly, different from existing two-stream networks, our FFCSN model enables us to cope with the temporal inconsistency between facial expressions and body motions for ADD. Moreover, we have also developed a new training approach for our FFCSN model by inducing meta learning and adversarial learning into the training process of our base model. As a result, our FFCSN model can be trained effectively even with only a handful of training samples. Extensive experiments show that the proposed FFCSN model achieves state-of-the-art results in both deception detection and emotion recognition.


\begin{thebibliography}{10}\itemsep=-1pt

\bibitem{abouelenien2017detecting}
M.~Abouelenien, V.~P{\'e}rez-Rosas, R.~Mihalcea, and M.~Burzo.
\newblock Detecting deceptive behavior via integration of discriminative
  features from multiple modalities.
\newblock {\em IEEE Trans. Information Forensics and Security},
  12(5):1042--1055, 2017.

\bibitem{chopra2005learning}
S.~Chopra, R.~Hadsell, and Y.~LeCun.
\newblock Learning a similarity metric discriminatively, with application to
  face verification.
\newblock In {\em CVPR}, pages 539--546, 2005.

\bibitem{depaulo2003cues}
B.~M. DePaulo, J.~J. Lindsay, B.~E. Malone, L.~Muhlenbruck, K.~Charlton, and
  H.~Cooper.
\newblock Cues to deception.
\newblock {\em Psychological Bulletin}, 129(1):74--118, 2003.

\bibitem{derksen2012control}
M.~Derksen.
\newblock Control and resistance in the psychology of lying.
\newblock {\em Theory \& Psychology}, 22(2):196--212, 2012.

\bibitem{feichtenhofer2016convolutional}
C.~Feichtenhofer, A.~Pinz, and A.~Zisserman.
\newblock Convolutional two-stream network fusion for video action recognition.
\newblock In {\em CVPR}, pages 1933--1941, 2016.

\bibitem{fornaciari2013automatic}
T.~Fornaciari and M.~Poesio.
\newblock Automatic deception detection in {Italian} court cases.
\newblock {\em Artificial Intelligence and Law}, 21(3):303--340, 2013.

\bibitem{gamer2014mind}
M.~Gamer.
\newblock Mind reading using neuroimaging: Is this the future of deception
  detection?
\newblock {\em European Psychologist}, 19(3):172, 2014.

\bibitem{gogate2017deep}
M.~Gogate, A.~Adeel, and A.~Hussain.
\newblock Deep learning driven multimodal fusion for automated deception
  detection.
\newblock In {\em IEEE Symposium Series on Computational Intelligence}, pages
  1--6, 2017.

\bibitem{goldberg2014word2vec}
Y.~Goldberg and O.~Levy.
\newblock word2vec explained: deriving {Mikolov} et al.'s negative-sampling
  word-embedding method.
\newblock {\em arXiv preprint arXiv:1402.3722}, 2014.

\bibitem{NIPS2014_5423}
I.~Goodfellow, J.~Pouget-Abadie, M.~Mirza, B.~Xu, D.~Warde-Farley, S.~Ozair,
  A.~Courville, and Y.~Bengio.
\newblock Generative adversarial nets.
\newblock In {\em Advances in Neural Information Processing Systems}, pages
  2672--2680, 2014.

\bibitem{graciarena2006combining}
M.~Graciarena, E.~Shriberg, A.~Stolcke, F.~Enos, J.~Hirschberg, and
  S.~Kajarekar.
\newblock Combining prosodic lexical and cepstral systems for deceptive speech
  detection.
\newblock In {\em ICASSP}, 2006.

\bibitem{he2016cvpr}
K.~He, X.~Zhang, S.~Ren, and J.~Sun.
\newblock Deep residual learning for image recognition.
\newblock In {\em CVPR}, pages 770--778, 2016.

\bibitem{hirschberg2005distinguishing}
J.~Hirschberg, S.~Benus, J.~M. Brenier, et~al.
\newblock Distinguishing deceptive from non-deceptive speech.
\newblock In {\em European Conference on Speech Communication and Technology},
  pages 1833--1836, 2005.

\bibitem{hoffer2015deep}
E.~Hoffer and N.~Ailon.
\newblock Deep metric learning using triplet network.
\newblock In {\em International Workshop on Similarity-Based Pattern
  Recognition}, pages 84--92, 2015.

\bibitem{howard2011acoustic}
D.~M. Howard and C.~Kirchh{\"u}bel.
\newblock Acoustic correlates of deceptive speech--an exploratory study.
\newblock In {\em International Conference on Engineering Psychology and
  Cognitive Ergonomics}, pages 28--37, 2011.

\bibitem{jaiswal2016truth}
M.~Jaiswal, S.~Tabibu, and R.~Bajpai.
\newblock The truth and nothing but the truth: Multimodal analysis for
  deception detection.
\newblock In {\em ICDM Workshops}, pages 938--943, 2016.

\bibitem{jiang2014predicting}
Y.-G. Jiang, B.~Xu, and X.~Xue.
\newblock Predicting emotions in user-generated videos.
\newblock In {\em AAAI}, volume~14, pages 73--79, 2014.

\bibitem{kozel2005detecting}
F.~A. Kozel, K.~A. Johnson, Q.~Mu, E.~L. Grenesko, S.~J. Laken, and M.~S.
  George.
\newblock Detecting deception using functional magnetic resonance imaging.
\newblock {\em Biological Psychiatry}, 58(8):605--613, 2005.

\bibitem{krishnamurthy2018deep}
G.~Krishnamurthy, N.~Majumder, S.~Poria, and E.~Cambria.
\newblock A deep learning approach for multimodal deception detection.
\newblock {\em arXiv preprint arXiv:1803.00344}, 2018.

\bibitem{levitan2016combining}
S.~I. Levitan, G.~An, M.~Ma, R.~Levitan, A.~Rosenberg, and J.~Hirschberg.
\newblock Combining acoustic-prosodic, lexical, and phonotactic features for
  automatic deception detection.
\newblock In {\em INTERSPEECH}, pages 2006--2010, 2016.

\bibitem{levitan2015cross}
S.~I. Levitan, G.~An, M.~Wang, G.~Mendels, J.~Hirschberg, M.~Levine, and
  A.~Rosenberg.
\newblock Cross-cultural production and detection of deception from speech.
\newblock In {\em ACM Workshop on Multimodal Deception Detection}, pages 1--8,
  2015.

\bibitem{li2017adversarial}
J.~Li, W.~Monroe, T.~Shi, S.~Jean, A.~Ritter, and D.~Jurafsky.
\newblock Adversarial learning for neural dialogue generation.
\newblock {\em arXiv preprint arXiv:1701.06547}, 2017.

\bibitem{LC10}
P.~Lucey, J.~F. Cohn, T.~Kanade, J.~Saragih, Z.~Ambadar, and I.~Matthews.
\newblock The extended {Cohn-Kanad} dataset ({CK+}): A complete dataset for
  action unit and emotion-specified expression.
\newblock In {\em CVPR Workshops}, pages 94--101, 2010.

\bibitem{mcduff2017large}
D.~McDuff and M.~Soleymani.
\newblock Large-scale affective content analysis: Combining media content
  features and facial reactions.
\newblock In {\em IEEE International Conference on Automatic Face \& Gesture
  Recognition}, pages 339--345, 2017.

\bibitem{michael2010motion}
N.~Michael, M.~Dilsizian, D.~Metaxas, and J.~K. Burgoon.
\newblock Motion profiles for deception detection using visual cues.
\newblock In {\em ECCV}, pages 462--475, 2010.

\bibitem{ogawa2017human}
T.~Ogawa, Y.~Yamaguchi, S.~Asamizu, and M.~Haseyama.
\newblock Human-centered video feature selection via {mRMR-SCMMCCA} for
  preference extraction.
\newblock {\em IEICE Trans. Information and Systems}, 100(2):409--412, 2017.

\bibitem{owayjan2012design}
M.~Owayjan, A.~Kashour, N.~Al~Haddad, M.~Fadel, and G.~Al~Souki.
\newblock The design and development of a lie detection system using facial
  micro-expressions.
\newblock In {\em International Conference on Advances in Computational Tools
  for Engineering Applications}, pages 33--38, 2012.

\bibitem{pang2015deep}
L.~Pang, S.~Zhu, and C.-W. Ngo.
\newblock Deep multimodal learning for affective analysis and retrieval.
\newblock {\em IEEE Trans. Multimedia}, 17(11):2008--2020, 2015.

\bibitem{peng2016multi}
X.~Peng and C.~Schmid.
\newblock Multi-region two-stream {R-CNN} for action detection.
\newblock In {\em ECCV}, pages 744--759, 2016.

\bibitem{perez2015deception}
V.~P{\'e}rez-Rosas, M.~Abouelenien, R.~Mihalcea, and M.~Burzo.
\newblock Deception detection using real-life trial data.
\newblock In {\em International Conference on Multimodal Interaction}, pages
  59--66, 2015.

\bibitem{perez2015verbal}
V.~P{\'e}rez-Rosas, M.~Abouelenien, R.~Mihalcea, Y.~Xiao, C.~Linton, and
  M.~Burzo.
\newblock Verbal and nonverbal clues for real-life deception detection.
\newblock In {\em EMNLP}, pages 2336--2346, 2015.

\bibitem{NIPS2015_5638}
S.~Ren, K.~He, R.~Girshick, and J.~Sun.
\newblock Faster {R-CNN}: Towards real-time object detection with region
  proposal networks.
\newblock In {\em Advances in Neural Information Processing Systems}, pages
  91--99, 2015.

\bibitem{santoro2017simple}
A.~Santoro, D.~Raposo, D.~G. Barrett, M.~Malinowski, R.~Pascanu, P.~Battaglia,
  and T.~Lillicrap.
\newblock A simple neural network module for relational reasoning.
\newblock In {\em Advances in Neural Information Processing Systems}, pages
  4967--4976, 2017.

\bibitem{simonyan2014two}
K.~Simonyan and A.~Zisserman.
\newblock Two-stream convolutional networks for action recognition in videos.
\newblock In {\em Advances in Neural Information Processing Systems}, pages
  568--576, 2014.

\bibitem{soomro2012ucf101}
K.~Soomro, A.~R. Zamir, and M.~Shah.
\newblock {UCF101}: A dataset of 101 human actions classes from videos in the
  wild.
\newblock {\em arXiv preprint arXiv:1212.0402}, 2012.

\bibitem{vrij2001detecting}
A.~Vrij.
\newblock {\em Detecting Lies and Deceit: The Psychology of Lying and
  Implications for Professional Practice}.
\newblock Wiley Series on the Psychology of Crime, Policing and Law. Wiley,
  Hoboken, NJ, USA, 2001.

\bibitem{vrij2006detecting}
A.~Vrij, R.~Fisher, S.~Mann, and S.~Leal.
\newblock Detecting deception by manipulating cognitive load.
\newblock {\em Trends in cognitive sciences}, 10(4):141--142, 2006.

\bibitem{Wang2013iccv}
H.~Wang and C.~Schmid.
\newblock Action recognition with improved trajectories.
\newblock In {\em ICCV}, pages 3551--3558, 2013.

\bibitem{wang2016temporal}
L.~Wang, Y.~Xiong, Z.~Wang, Y.~Qiao, D.~Lin, X.~Tang, and L.~Van~Gool.
\newblock Temporal segment networks: Towards good practices for deep action
  recognition.
\newblock In {\em ECCV}, pages 20--36, 2016.

\bibitem{wang2015video}
S.~Wang and Q.~Ji.
\newblock Video affective content analysis: a survey of state of the art
  methods.
\newblock {\em IEEE Trans. Affective Computing}, 6(4):410--430, 2015.

\bibitem{wei2018deep}
X.-S. Wei, C.-L. Zhang, H.~Zhang, and J.~Wu.
\newblock Deep bimodal regression of apparent personality traits from short
  video sequences.
\newblock {\em IEEE Trans. Affective Computing}, 9(3):303--315, 2018.

\bibitem{wu2018deception}
Z.~Wu, B.~Singh, L.~S. Davis, and V.~S. Subrahmanian.
\newblock Deception detection in videos.
\newblock In {\em AAAI}, pages 1695--1702, 2018.

\bibitem{xia2007deception}
F.~Xia, H.~Wang, and J.~Huang.
\newblock Deception detection via blob motion pattern analysis.
\newblock In {\em International Conference on Affective Computing and
  Intelligent Interaction}, pages 727--728, 2007.

\bibitem{xie2018comparator}
W.~Xie, L.~Shen, and A.~Zisserman.
\newblock Comparator networks.
\newblock In {\em ECCV}, pages 811--826, 2018.

\bibitem{xu2016video}
B.~Xu, Y.~Fu, Y.-G. Jiang, B.~Li, and L.~Sigal.
\newblock Video emotion recognition with transferred deep feature encodings.
\newblock In {\em ICMR}, pages 15--22, 2016.

\bibitem{xu2018heterogeneous}
B.~Xu, Y.~Fu, Y.-G. Jiang, B.~Li, and L.~Sigal.
\newblock Heterogeneous knowledge transfer in video emotion recognition,
  attribution and summarization.
\newblock {\em IEEE Trans. Affective Computing}, 9(2):255--270, 2018.

\bibitem{yang2016wider}
S.~Yang, P.~Luo, C.-C. Loy, and X.~Tang.
\newblock {WIDER} {FACE}: A face detection benchmark.
\newblock In {\em CVPR}, pages 5525--5533, 2016.

\bibitem{zhang2016bimodal}
C.-L. Zhang, H.~Zhang, X.-S. Wei, and J.~Wu.
\newblock Deep bimodal regression for apparent personality analysis.
\newblock In {\em ECCV}, pages 311--324, 2016.

\bibitem{zhang2018recognition}
H.~Zhang and M.~Xu.
\newblock Recognition of emotions in user-generated videos with kernelized
  features.
\newblock {\em IEEE Trans. Multimedia}, 20(10):2824--2835, 2018.

\bibitem{zhang2007real}
Z.~Zhang, V.~Singh, T.~E. Slowe, S.~Tulyakov, and V.~Govindaraju.
\newblock Real-time automatic deceit detection from involuntary facial
  expressions.
\newblock In {\em CVPR}, pages 1--6, 2007.

\bibitem{Zhou_2018_ECCV}
B.~Zhou, A.~Andonian, A.~Oliva, and A.~Torralba.
\newblock Temporal relational reasoning in videos.
\newblock In {\em ECCV}, pages 803--818, 2018.

\bibitem{Zolfaghari_2018_ECCV}
M.~Zolfaghari, K.~Singh, and T.~Brox.
\newblock {ECO}: Efficient convolutional network for online video
  understanding.
\newblock In {\em ECCV}, pages 695--712, 2018.

\end{thebibliography}
\end{document}